\documentclass[conference,10pt]{IEEEtran}

\usepackage[utf8]{inputenc}
\usepackage[T1]{fontenc}
\usepackage{textcomp}
\usepackage{fixltx2e}
\usepackage{multirow}
\usepackage{algorithm}
\usepackage{algpseudocode}
\usepackage[pdftex]{graphicx}
\usepackage{graphicx}
\DeclareGraphicsExtensions{.pdf,.png,.jpg,.eps}
\usepackage[caption=false,font=footnotesize]{subfig}
\usepackage{booktabs}
\usepackage{url}
\usepackage{amsmath}
\usepackage{amssymb}
\usepackage{amsthm}
\usepackage[capitalise,noabbrev]{cleveref}
\usepackage[textwidth=1cm]{todonotes}
\usepackage{xspace}
\usepackage{booktabs}
\usepackage[textwidth=1cm]{todonotes}
\usepackage[american]{babel}
\usepackage[color]{changebar}
\usepackage{textgreek}
\usepackage{enumitem}
\usepackage[bottom]{footmisc}
\usepackage{stfloats}
\usepackage{mathrsfs}
\usepackage[square, numbers]{natbib}

\usepackage[load-configurations=binary,detect-all,binary-units=true,range-phrase=--,per-mode=symbol,range-units=single,list-units=single]{siunitx}
\DeclareSIUnit\dBm{dBm}
\DeclareSIUnit\dB{dB}

\RequirePackage{xstring}
\RequirePackage{xparse}
\RequirePackage[]{acro}
\NewDocumentCommand\acrodef{mO{#1}mG{}}{\DeclareAcronym{#1}{short={#2}, long={#3}, #4}}
\acrodef{ATIS}{advanced traveler information system}{short-plural=}
\acrodef{AVCS}{advanced vehicle control system}{short-plural=}
\acrodef{ATMS}{advanced traffic management system}{short-plural=}
\acrodef{CPSS}{cyber-physical social system}{short-plural=}
\acrodef{DQN}{deep Q-network}
\acrodef{DRL}{deep reinforcement learning}
\acrodef{GIN-E}{graph isomorphism network with edge features}
\acrodef{GNN}{graph neural network}
\acrodef{IoT}{internet of things}
\acrodef{ITS}{intelligent transportation system}{short-plural=}
\acrodef{RL}{reinforcement learning}
\acrodef{V2X}{vehicle-to-everything}

\newcommand{\hv}{\mathbf{h}}
\newcommand{\pv}{\mathbf{p}}
\newcommand{\Pv}{\mathbf{P}}
\newcommand{\sv}{\mathbf{s}}
\newcommand{\Hv}{\mathbf{H}}

\newcommand{\Gc}{\mathcal{G}}
\newcommand{\Ec}{\mathcal{E}}
\newcommand{\Rc}{\mathcal{R}}
\newcommand{\Sc}{\mathcal{S}}
\newcommand{\Vc}{\mathcal{V}}
\usepackage{lineno}
\IEEEoverridecommandlockouts
\begin{document}

\title{Personalized and Context-aware Route Planning \\for Edge-assisted Vehicles}

\author{
\IEEEauthorblockN{Dinesh Cyril Selvaraj\IEEEauthorrefmark{1}, Falko Dressler\IEEEauthorrefmark{2}, Carla Fabiana Chiasserini\IEEEauthorrefmark{1}}
\IEEEauthorblockA{\IEEEauthorrefmark{1}CARS@Polito, Politecnico di Torino, Italy}
\IEEEauthorblockA{\IEEEauthorrefmark{2}School of Electrical Engineering and Computer Science, TU Berlin, Germany}
}
\maketitle

\begin{abstract}
Conventional route planning services typically offer the same routes to all drivers, focusing primarily on a few standardized factors such as travel distance or time, overlooking individual driver preferences.
With the inception of autonomous vehicles expected in the coming years, where vehicles will rely on routes decided by such planners, there arises a need to incorporate the specific preferences of each driver, ensuring personalized navigation experiences.
In this work, we propose a novel approach based on \acp{GNN} and \ac{DRL}, aimed at customizing routes to suit individual preferences.
By analyzing the historical trajectories of individual drivers, we classify their driving behavior and associate it with relevant road attributes as indicators of driver preferences.
The \ac{GNN} is capable of representing the road network as graph-structured data effectively, while \ac{DRL} is capable of making decisions utilizing reward mechanisms to optimize route selection with factors such as travel costs, congestion level, and driver satisfaction.
We evaluate our proposed GNN-based DRL framework using a real-world road network and demonstrate its ability to accommodate driver preferences, offering a range of route options tailored to  individual drivers. 
\textcolor{black}{The results indicate that our framework can select routes that accommodate driver's preferences with up to a $17$\% improvement compared to a generic route planner, and reduce the travel time by $33$\% (afternoon) and $46$\% (evening) relatively  to the shortest distance-based approach.}
\end{abstract}


%

\section{Introduction}

Over the past few decades, the number of people migrating to urban areas has been on the rise, reaching now more than half of the  world population and expected to 
 double by 2050 \cite{UNReport_population}.
To address the challenges posed by rapid urbanization, cities are employing technological solutions across multiple domains to ensure their effective operation.
The adoption of technological solutions in the transport domain is collectively referred to as \acp{ITS}.
Cooperative ITS solutions \cite{EU_CCAM_2016}, in particular, are becoming mainstream, with a focus on sharing information with other vehicles and road infrastructures to enable advanced vehicular applications. 
Specifically, such cooperative solutions foster the development of data-driven models wherein a driver can make informed decisions based on real-time traffic data to enhance road safety, traffic efficiency, and driving experiences \cite{sommer2014vehicular,autili2021cooperative,dinesh-journal}.

Vehicle route planning is one such application that can benefit from utilizing information shared between vehicles and road infrastructures \cite{liu2019intelligent, schoenberg2023reducing}.
Effective route planning can improve people's lifestyles as they spend a substantial amount of time commuting, and also impact significantly  relevant factors like people's health, air quality, and  fuel consumption   \cite{inrix_scorecard_nodate, HEIReport}.
In addition to traffic efficiency, it is crucial to incorporate user preferences to enhance the driving experience -- creating next generation \acp{CPSS} \cite{dressler2018cpss}.
Current routing service providers often lack personalized route planning options apart from generic options like toll avoidance, highlighting the need for facing this challenge. 
Previous approaches    proposed in the literature typically  leverage only historical driving data \cite{dai2015personalized}, or focus on identifying risk factors associated with specific road segments \cite{abdelrahman2020crowdsensing-based} such as areas with potential criminal activities \cite{de-souza2020safe}.

Unlike existing works, to effectively address  personalized route planning,  we propose a robust, context-aware route planner that considers both traffic-related information and driver preferences to determine customized route for the driver.
We employ a novel framework that combines two machine learning paradigms: \ac{GNN} to represent the road network as graph-structured data, and \ac{DRL} for its decision-making capabilities in highly dynamic environments.
The integration of these approaches fosters the development of a framework that can make informed decisions accounting for dynamic traffic-related information and individual preferences.
To the best of our knowledge, we are the first to combine \acp{GNN} with \ac{DRL} to address personalized and context-aware route planning systems.
We evaluate the performance of the proposed framework using a realistic road traffic network \cite{codeca2017luxembourg} through the open-source traffic simulator SUMO \cite{behrisch2011sumo}.

Our main contributions are thus  as follows:
\begin{itemize}
\item[{\em (i)}] We present a novel framework that integrates \acp{GNN} with \ac{DRL}, leveraging real-time traffic information of the road network represented as graph-structured data.
With the help of carefully curated reward components, the framework is capable of providing tailored and time-efficient routes for drivers.

\item[{\em (ii)}] We determine drivers' preferences by analyzing their driving behavior.
In particular, we classify drivers' behavior into aggressive and normal based on historical driving data.
Subsequently, we identify individual drivers' preferences that correspond to specific road attributes observed during normal driving behavior.

\item[{\em (iii)}]\textcolor{black}{ We introduce a flexible training approach to handle various driving preferences where a generic model is trained considering the generic traffic routing options.
 Subsequently, multiple models are trained on this generic model, each tailored to specific driving preferences, incorporating driver's satisfaction rewards to provide personalized routes. 
This also allows our framework to accommodate the multiple variations of evolving driver's preferences. 
Experimental results reveal that  models trained to prioritize driver's preferences can achieve up to a $17$\% improvement in selecting routes preferred by drivers when compared to the generic model,  while  reducing the travel time by up to $46$\%  relatively  to the shortest distance-based approach.}
\end{itemize}


%

\section{Related Work\label{sec:relwrk}}

Vehicle route planning is a widely explored  topic, and numerous methodologies have been proposed to determine efficient routes to reach a destination considering multiple factors, including travel time, distance, fuel consumption, charging time of EV, and  environmental impact \cite{sarker2020data-driven, guidoni2020vehicular, pan2013proactive, schoenberg2023reducing}.
Early route optimization systems utilized k-shortest path (kSP) algorithms \cite{yen1970algorithm}, primarily relying on static features like travel distance and employing Dijkstra's algorithm \cite{dijkstra1959note} or the Bellman-Ford algorithm \cite{bellman1958routing} to determine the shortest path. However, the shortest path in terms of distance does not necessarily equate to the fastest route due to dynamic factors like traffic congestion and road conditions.

Recent technological advancements and the integration of real-time data have led to the development of new methodologies, particularly variants of kSP algorithms \cite{pan2013proactive}, accounting for dynamic factors such as travel time, alternative vehicle routes, and real-time traffic conditions.
Furthermore, optimal velocity profiles have been proposed for vehicles to align with traffic flow and traffic light management strategies to mitigate frequent braking \cite{ding2017greenplanner, koukoumidis2011signalguru}. 
These methodologies, however, often prioritize factors like travel time and distance, overlooking human preferences, whereas few studies \cite{li2017pare, quercia2014shortest, dai2015personalized} account for human preferences in route optimization tailored to drivers' preferences.

Few studies \cite{quercia2014shortest, abdelrahman2020crowdsensing-based} leveraged  crowdsourced data for route planning: \citet{quercia2014shortest} suggested routes based on noise level and scenic views, while \citet{abdelrahman2020crowdsensing-based} accounted for road quality and individual preferences. \citet{li2017pare} utilized drivers' historical data to identify frequently traveled routes and make suggestions that incorporate familiar landmarks along the path while minimizing the number of route segments through dynamic programming techniques. Similarly, \citet{dai2015personalized} proposed taxi route planning based on historical driver data, prioritizing travel time and fuel efficiency, while \citet{de-souza2020safe} prioritized safe areas using historical crime data along with sporadic traffic reports for travel time estimation. However, neither of these approaches utilized real-time traffic information, which could lead to inaccurate travel time estimation, particularly in non-recurring traffic situations.
Similarly, \citet{schoenberg2023reducing} tackled route optimization  for EV charging.
One major disadvantage of the proposed approaches is their inability to provide routes in near real-time manner, as it takes time to identify the solutions using Pareto front-based optimization techniques. 
The computation time comparison taken by \ac{RL}-based route recommendation systems with respect to Pareto optimal algorithms highlights that RL-based systems take negligible time to suggest  routes compared to other approaches  \cite{sarker2020data-driven}. 

There are a few studies utilizing RL to find a suitable path between the source and destinations based on different factors, which are represented as rewards \cite{sarker2020data-driven, panov2018grid, zhao2021hybrid, lei2022solve}.
\citet{sarker2020data-driven} proposed a methodology to find a suitable path considering fuel consumption, travel time, air quality, travel distance as rewards and further allow for tuning the reward components according to human preferences.
A Q-learning based algorithm for finding the shortest path in a grid-based environment is employed in \cite{panov2018grid}, which is less intricate than real-world networks used for framework validation.
In contrast to such methodology using graph-structured data to represent states, our  approach can address generic, possibly more complex, scenarios.

Graph-structured data indeed facilitate efficient information exchange between nodes and maintaining up-to-date information on all available paths and their corresponding traffic conditions.
Although \ac{GNN} and \ac{RL} techniques are utilized in \cite{zhao2021hybrid, lei2022solve}, these solutions primarily focus on the traveling salesman problem with an assumption of relatively stable traffic conditions.
However, the vehicle route planning presents an additional challenge which aims to identify the optimal route between source and destination considering highly dynamic factors.
We address such vehicle routing problem by considering factors such as travel time, congestion level, and driver satisfaction, employing both \ac{GNN} and \ac{DRL}.
By utilizing such a framework, real-time traffic information is effectively integrated into the graph structure, enabling the provision of near real-time routing solutions compared to traditional algorithms.

%

\section{Preliminaries\label{sec:background}}

In this work, we integrated \aclp{GNN} and \acl{RL} to harness the advantages of representing road networks as graph-structured data and to leverage the decision-making capabilities of \ac{RL} in providing a scalable and time-efficient personalized route planner.
\Acp{GNN} play a role in encoding road network-related information as graph structures, effectively capturing relationships and dependencies between nodes. 
They capture both local and global information by passing messages between neighboring nodes.
In the context of the road network, nodes represent intersections and edges represent road connections between nodes, each associated with features to effectively represent the road network for the route planner task at hand.
The reinforcement learning agent considers such graph-structured data representing the road network attributes as environment states to make decisions based on the graph-structured data using the GNN-based DRL policy network.

\subsection{Graph neural networks}

A graph neural network is a type of neural network that deals with graph-structured data.
Graph-structured data contains a set of entities represented as nodes, and connections between them are represented as edges, making them suitable for representing complex relationships that capture dependencies and interactions between nodes.
Such representation of data as graphs has gained significant attention in recent years due to its effectiveness in various domains \cite{zhou2020graph}, such as recommendation systems, social networks, and molecular biology.
\Acp{GNN} can be applied to various tasks: node-level classification within a graph, edge-level prediction to identify if two nodes share an edge, and graph-level prediction to predict the property of an entire graph. In this work, we utilize the graph-level task to predict the global representation of the entire graph.

The core operation of the GNN is the message passing process 
that takes place between nodes and their neighbors.
At each iteration, each node receives information from its neighbors' feature vectors and aggregate the messages received from its neighbors.
The message passing and aggregation operation can be defined as
\begin{equation}\label{eq:mp}
\begin{split}
m^{(l)}_{i} = \sum_{j \in N(i)} M(h_{i}^{(l)}, h_{j}^{(l)}, e_{ij})
\end{split}
\end{equation}
where $M(\cdot)$ represents the neural network as a function approximator, $h_{i}^{(l)}, h_{j}^{(l)}$ represent the features vectors of nodes $i$ and $j$ (resp.) at iteration $l$, $N(i)$ represents the set of neighboring nodes of node $i$, and $e_{ij}$ represents the  features of the edge between nodes $i$ and $j$.
Later, each node updates its feature vector based on the aggregated messages and its own previous feature vector using the approximated function approximator. 

After repeating the  message passing process for a certain number of iterations, a readout function is employed to aggregate all final node representations into a graph-level representation as the output.
The output of the readout function can be utilized for the downstream tasks such as graph classification/regression.

In this work, we make use of the \ac{GIN-E}  \cite{hu2019strategies} architecture, which is a variant of the Graph Isomorphism Network (GIN) \cite{xu2019how}.
It extends the GIN model by incorporating edge features into the graph representation.
The main advantage of using  the \ac{GIN-E} variant is the capability to associate both node features and edge features when updating node representations during the message passing process.
Such variation is crucial for the use case discussed in this work, as the edge features play a vital role in determining the available routing options.

\subsection{Reinforcement learning}

Reinforcement learning is a type of machine learning paradigm wherein an RL agent iteratively interacts with the environment, taking action and receiving feedback in the form of reward. 
This interaction is typically formalized as a discounted Markov Decision Process (MDP), denoted as a tuple $(\mathcal{S}, \mathcal{A}, \mathcal{P}, \mathcal{R}, \gamma)$, representing the state space, action space, state transition probability, reward, and discount factor, respectively. Moreover, the agent's policy $\pi$ represents its decision-making strategy aimed at optimizing the expected discounted reward.

Q-learning \cite{watkins1992qlearning} is an RL algorithm designed to find the optimal policy $\pi$, given the Markov decision process.
The idea behind the Q-Learning is to iteratively interact with the environment to learn the optimal action-value function $Q(s, a)$ representing the expected cumulative future rewards by taking action $a$ in state $s$ assuming the agent follows the current policy $\pi$ thereafter.
During the learning process, the action-value function $Q(s, a)$ is updated using the Bellman equation 
\begin{equation}\label{eq:qvalueFunction}
\begin{split}
Q(s, a) \longleftarrow Q(s, a) + \alpha(r(s, a) + \\\gamma\max_{a'}Q(s', a') - Q(s, a))
\end{split}
\end{equation}
where $Q(s, a)$ represents the current Q-value for state $s$ and action $a$, $\alpha$ represents the learning rate, $r(s, a)$ represents the immediate reward after taking action $a$ in state $s$, $s'$ represents the resulting state after taking action $a$, and $a'$ indicates the possible actions to take in the resulting state $s'$.

As the basic Q-Learning algorithm encounters difficulties in learning an optimal policy with high-dimensional state and action spaces, \ac{DRL} employs a deep neural network as function approximator, enabling it to manage complex decision-making tasks. 
In this study,  the deep neural network adaptation of the Q-Learning algorithm, the \ac{DQN} \cite{mnih2013playing}, is used to learn the optimal policy.

%

\section{Personalized and Context-aware Route Planning\label{sec:methodology}}


In this section, we present our framework for  personalized and context-aware route planning.
To achieve the desired outcome, we have to address two key questions:
{\em (1) how to determine  the desired path for the given driver}, and
{\em (2) how to provide a route that meets driver's individual preferences,
without compromising the traffic efficiency factors such as travel time and traffic congestion level}.
We tackle these challenges using two main components:
\begin{enumerate}
    \item a driver behavior classifier, which links the driver's behaviors with the road network attributes using historical driver data, which helps to identify the individual preferences of the driver based on their driving pattern, and
    \item an approach that combines the benefits of GNNs and DRL to learn an optimal policy for personalized end-to-end route selection, with the goal to maximize driver's satisfaction while minimizing travel time and mitigating global congestion on the road network. 
\end{enumerate}

The integrated framework combines graph-based learning and RL, updating both model parameters simultaneously to create a robust, adaptive system capable of making decisions based on real-time data from the graph-structured road network.
Further, to enhance flexibility and accommodate multiple driver preferences, a generic model is initially trained on traditional traffic-related factors such as travel time and traffic congestion.
Subsequently, multiple driver preference-oriented models are trained based on this generic model, each with satisfaction rewards tailored to different drivers' preferences.
The framework is designed to receive routing requests from vehicles in the form of a tuple {\tt <source, destination>}, aiming to identify routes that connect them while prioritizing factors such as driver's satisfaction, global congestion, and travel time.
It is important to note that we approximate the requested route as a junction closest to the driver’s starting point (referred to as the source junction) to the destination junction closest to the desired destination point.

\subsection{System model\label{sub:sysmodel}}


We represent the road network as a directed graph $\Gc {=} (\Vc, \Ec)$, where the vertex set $\Vc$ represents the roads junctions and the edge set $\Ec$ models the roads connecting two road junctions.
As an example, edge $e_{ij} {=} (v_i, v_j)$ corresponds to the road connecting the junctions $i$ and $j$.
A route $\Rc{=} <e_{ij}, e_{jk},\ldots> $ is a list of edges connecting a sequence of road junctions, where consecutive edges share a vertex. Each edge carries a set of attributes that represent the current state of the network.
\Cref{tab:attributes} summarizes the road network attributes between vertices $i$ and $j$ at any given time $t$.

\begin{table}
\centering
\caption{Road network attributes}
\label{tab:attributes}
\begin{tabular}{ll}
    \toprule      
    Symbols & Description \\
    \midrule
    $\rho_{ij}(t)$ & \begin{tabular}{@{}l@{}} Occupancy percentage, percentage of road length \\ occupied by the vehicles on the edge $e_{ij}$\end{tabular}\\
    $\mathcal{U}_{ij}(t)$ & Road usage, number of vehicles on the edge $e_{ij}$ \\
    $\dot{\mathcal{U}}_{ij}(t)$ & Road speed, average speed of vehicles on the edge $e_{ij}$ \\
    $\xi_{ij}(t)$ & Travel time, time taken to travel through the edge $e_{ij}$ \\
    $\psi_{ij}$ & Number of lanes on the edge $e_{ij}$\\
    $\chi_{ij}$ & Road length of the edge $e_{ij}$ \\
    $\zeta_{ij}$ & Road type, structure of the road represented by the edge $e_{ij}$ \\
    $\nu_{ij}$ & \begin{tabular}{@{}l@{}} Road complexity, complex road junctions\\ such as roundabouts at the vertex $j$ \end{tabular}\\
    $\mathcal{F}_{ij}(t)$ & \begin{tabular}{@{}l@{}}Future road usage, number of vehicles \\that are expected to travel on the road $e_{ij}$ \end{tabular}\\ 
    \bottomrule 
\end{tabular}
\end{table}


The vehicle-related features are extracted from historical driving data to classify the driver's behavior and provide personalized routing solutions that align with their preferences.
\Cref{tab:veh_features} summarizes the features utilized to classify and associate the driver's  behavior with the road attributes.

\begin{table}
\centering
\caption{Vehicle-related features to classify driver behavior}
\label{tab:veh_features}
\begin{tabular}{ll}
    \toprule      
    Symbols & Description  \\
    \midrule
    $\tau$ & Timestamp, denoting the time of the recorded data\\
    $\lambda$ & Longitudinal position \\
    $\phi$ & Lateral position \\
    $\ddot{x}$ & Longitudinal vehicle acceleration \\
    $\ddot{y}$ & Lateral vehicle acceleration \\
    $\vartheta$ & Time headway \\
    $\dot{x}$ & Longitudinal velocity \\
    \bottomrule  
\end{tabular}
\end{table}


\subsection{Characterizing the driver's behavior and preferences \label{sec:DBC}}

In the context of route planning, there are several approaches to factoring in the driver's preferences to provide personalized routes for drivers \cite{dai2015personalized, abdelrahman2020crowdsensing-based}.
In our study, we analyze the driver’s behavior by considering both static and dynamic environmental attributes, to identify their preferred route. 

We build upon the work by \citet{takahashi2015examination}, which also divides the environmental attributes into static and dynamic.
The static elements comprise the information related to the physical topology of the road, e.g., road length and number of lanes.
The dynamic elements represent temporal attributes such as the current traffic situation, accidents, or weather.  
Among them, we select four major attributes to represent the road environmental vector $\hv_{ij}(t)$, which contains information about both the physical road topology and the dynamic information associated with any given edge $e_{ij}$.
The road environmental vector is defined as \[\hv_{ij}(t){\triangleq} \{\zeta_{ij}, \psi_{ij}, \nu_{ij}, \kappa_{ij}(t)\},\] where the vector components are as defined above.

The road environment vector conveys information about the road or edges, including factors such as road type $\zeta$, number of lanes $\psi$, road complexity $\nu$, and current traffic conditions $\kappa(t) {\in} [\text{low}, \text{high}]$.
To match these attributes with individual preferences, we study how drivers behave on roads with specific attribute combinations, gaining insights into their preferences and tendencies.
To understand driver preferences, we leverage the driver's historical data and associate them with the corresponding road environmental vector $\hv$ to define the driving preference vector $\pv$.
Subsequently, the road characteristics associated with the normal driver's behavior are identified as the preferred attributes for travel.

\begin{figure}
\center
\includegraphics[width=0.9\columnwidth]{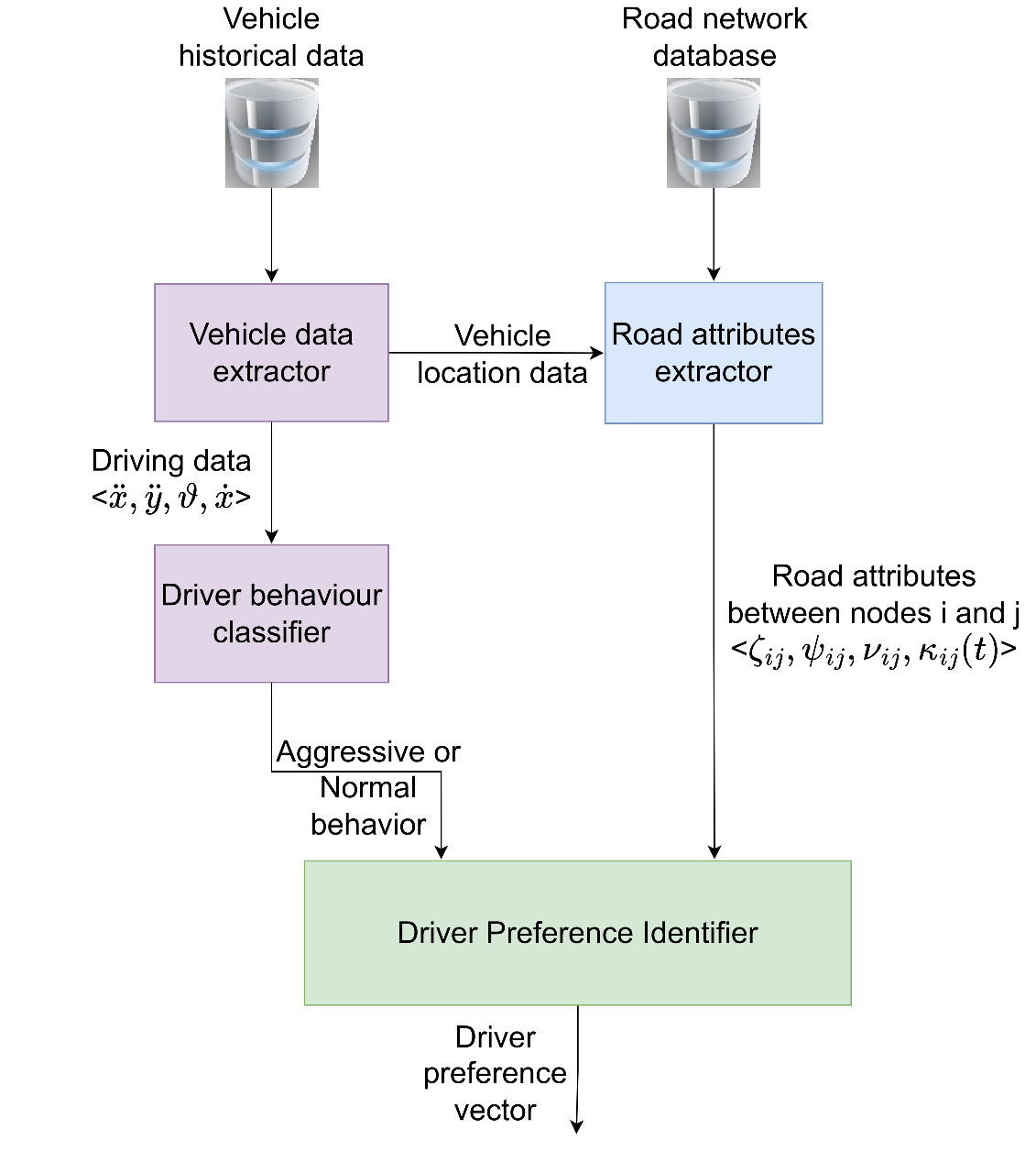}
\caption{An overview of the driver's preference identifier.\label{fig:drivpref}}
\end{figure}

\Cref{fig:drivpref} illustrates the process of identifying the driver's  preference vector.
Initially, we extract edge information from the road network database.
For each edge, we utilize the (${\ddot{x}, \ddot{y}, \vartheta,\dot{x}}$) parameters to classify the driver behavior $B_e$.
Once the driving behavior is determined, the driver preference identifier utilizes the road attributes vector $\Hv$ for each edge and associates them with the driving behavior.
Subsequently, the preference identifier associates the road attributes vector with the frequent normal driving behavior, considering it as the driver's preference vector $\pv$.

\subsection{GNN-based DRL agent\label{sec:GNNDRL}}

We now describe our GNN-based DRL framework.
We follow a similar approach as described by \citet{almasan2022deep}, which addresses the optical transport network routing use case.
In a nutshell, the GNN component is used for modeling the topology and the attributes of the road network as a directed graph.
The latter is then used as environment for the DRL agent. 
As mentioned before, we leverage the \ac{GIN-E} architecture \cite{hu2019strategies} -- a variant of GNN that facilitates the information exchange between the graph vertices, encompassing both node and edge features.
The DRL counterpart utilizes a DQN-based algorithm \cite{mnih2013playing, almasan2022deep}, which takes the graph-structured network state as input and predicts an action to determine the route.
The DRL uses the Q-value estimate provided by the GNN, based on the current network state and learns to minimize the mean-squared Bellman error (\cref{eq:qvalueFunction}). 





\textbf{\emph{State space:}}
The DRL state, $\sv(t) {\in} \Sc$, is the network state, defined by the edge-level features between vertices $i$ and $j$ at any given time $t$ (see Sec.\,\ref{sub:sysmodel}), i.e.,   
\[\sv(t){=}\{\rho_{ij}(t), \mathcal{U}_{ij}(t), \dot{\mathcal{U}}_{ij}(t), \xi_{ij}(t), \psi_{ij}, \chi_{ij}, \zeta_{ij}, \nu_{ij}, \mathcal{F}_{ij}(t)\}\,.\]  
Note that the network state described above comprises all the main parameters that we are aiming to include in optimizing the route selection process at the network topology level.

\textbf{\emph{Action space:}}
Upon receiving a route request at time $t$, the DRL selects an action, $a(t) {\in} \mathcal{A}$, with $\mathcal{A}$ denoting the set of end-to-end routes from the requested source to the destination.
Given the complexity of the network topology, the number of feasible route options from a source to a destination may be very high.  
%
Since estimating the  Q-values for every possible combination could significantly affect policy convergence, we limit the action space $\mathcal{A}$ to a set of $K$ candidate routes. 
The value of $K$ is a tunable hyperparameter aimed to achieve a good trade-off between the optimal route selection and the complexity to evaluate all the feasible routes (i.e., from source to destination) \cite{almasan2022deep}.
Based on the current travel time, we determine the $K$ candidate paths for each source-destination pair from a set of feasible routes.
%
The selected route is expressed in the network state by an attribute named "selected path", using an indicator function takes   1 for the selected edges and 0 for the others.
The DRL then estimates the Q-value based on the updated network states, learning to maximize cumulative rewards. 

\textbf{\emph{Reward components:}}
We define the reward, i.e., the numerical value received from the environment as feedback for the DRL agent’s action, as a multi-objective function.
It comprises three components:
(i) the driver's satisfaction, based on the history of the driver's behavior and preferences on the selected route;
(ii) the time taken to reach the destination; and
(iii) the global traffic flow to ensure smooth vehicular mobility in the considered area.
More formally, the reward can be written as:
\begin{equation}\label{eq:reward}
\begin{split}
r(\sv(t),a(t)) {=} \omega_{p} \cdot r_{p}(\sv(t),a(t)) + \omega_{t} \cdot r_{t}(\sv(t),a(t)) \\+ \phantom{\,}\omega_{f} \cdot r_{f}(\sv(t),a(t))
 \end{split}
\end{equation}
where 
$\omega_{p}$, $\omega_{t}$,  and $\omega_{f}$ are weighting coefficients and $r_{p}(\sv(t),a(t))$, $ r_{t}(\sv(t),a(t))$, and $r_{f}(\sv(t),a(t))$ are the reward components representing driver satisfaction, time-taken to reach the destination, and global traffic flow at time step $t$, respectively.
Each component allows values in the range [0,1].

\emph{Driver satisfaction}:
The driver satisfaction reward component has been formulated to provide a high reward for routing the vehicle along roads (edges) that are aligned with the driver's preference vector $\pv$ derived from the driver's behavior.
It is not always possible to identify a route that solely caters to the driver's routing choice; therefore, to encourage the DRL to choose a route with road attributes similar to the driver’s preferences, we utilize the weighted cosine similarity score, $S_c$, to model the reward component.
$S_c$ between the driver's preference $\pv$ and the road attributes $\hv$ is calculated for each edge along the selected route.
Moreover, it is possible to assign priority factors to specific elements of the driver's preference vectors, indicating which elements must correspond with road attributes during the route selection process.
Based on these priority factors, a list of values $S_p$ is generated: A value of 1 indicates that the preference vector element mentioned in the priority factor aligns with the corresponding road attribute. 
The similarity value ($S_e$) is computed for each edge in the selected route, and the mean of these values corresponds to the driver's satisfaction reward.
Finally, the driver's satisfaction reward is formulated as
\begin{eqnarray} \label{eq:prefReward}
S_e {=} \min(\text{mean}(S_p), S_c) \\
r_{p}(\sv(t),a(t)) {=} \text{mean}(S_e)\,.
\end{eqnarray}

\emph{Time to reach the destination}:
This reward is modeled using an error function $\text{erf}$ between the ideal time and the actual time to reach the destination from the source node.
The ideal time $T_i$ is computed based on the distance between each node pair from the source to destination with the speed limit specified between them.
Note that the ideal time does not account for the time to cross the junction or the waiting period at the traffic lights.
The actual time $T_a$, instead, is calculated based on the current average speed along the route, which includes the current traffic conditions as well.
The reward function is designed to provide the maximum reward if $T_a{=}T_i$; however, the function is also modeled to provide a lower reward even if they reach the destination before the ideal time, so as to discourage the driver from exceeding the speed limit.
\begin{eqnarray} \label{ttEqn}
&& r_{t}(\sv(t),a(t)){=}
\begin{cases} 1 {-} \text{erf}(\frac{T_i {-} T_a}{T_i}), T_i {\geq} T_a\\ 
              1 {-} \text{erf}(\frac{T_a {-} T_i}{T_i}), T_a {>} T_i \,.
\end{cases}
\end{eqnarray}

\emph{Global traffic flow}:
The occupancy percentage $\rho_{ij}$ for each edge $ij$ in the road network is obtained by calculating the length of vehicles on the road along with their minimum gap, divided by the road length.
The occupancy percentage represents the current global traffic flow.
To avoid congestion, the DRL is encouraged to route the vehicle through less crowded routes by providing a reward based on the global mean occupancy rate as 
\begin{equation}\label{eq:flwReward}
r_{f}(\sv(t),a(t)) {=} 1 {-} \text{mean}(\rho_{ij}) \,.
\end{equation}



\begin{figure}
    \centering
    \includegraphics[width=1\columnwidth]{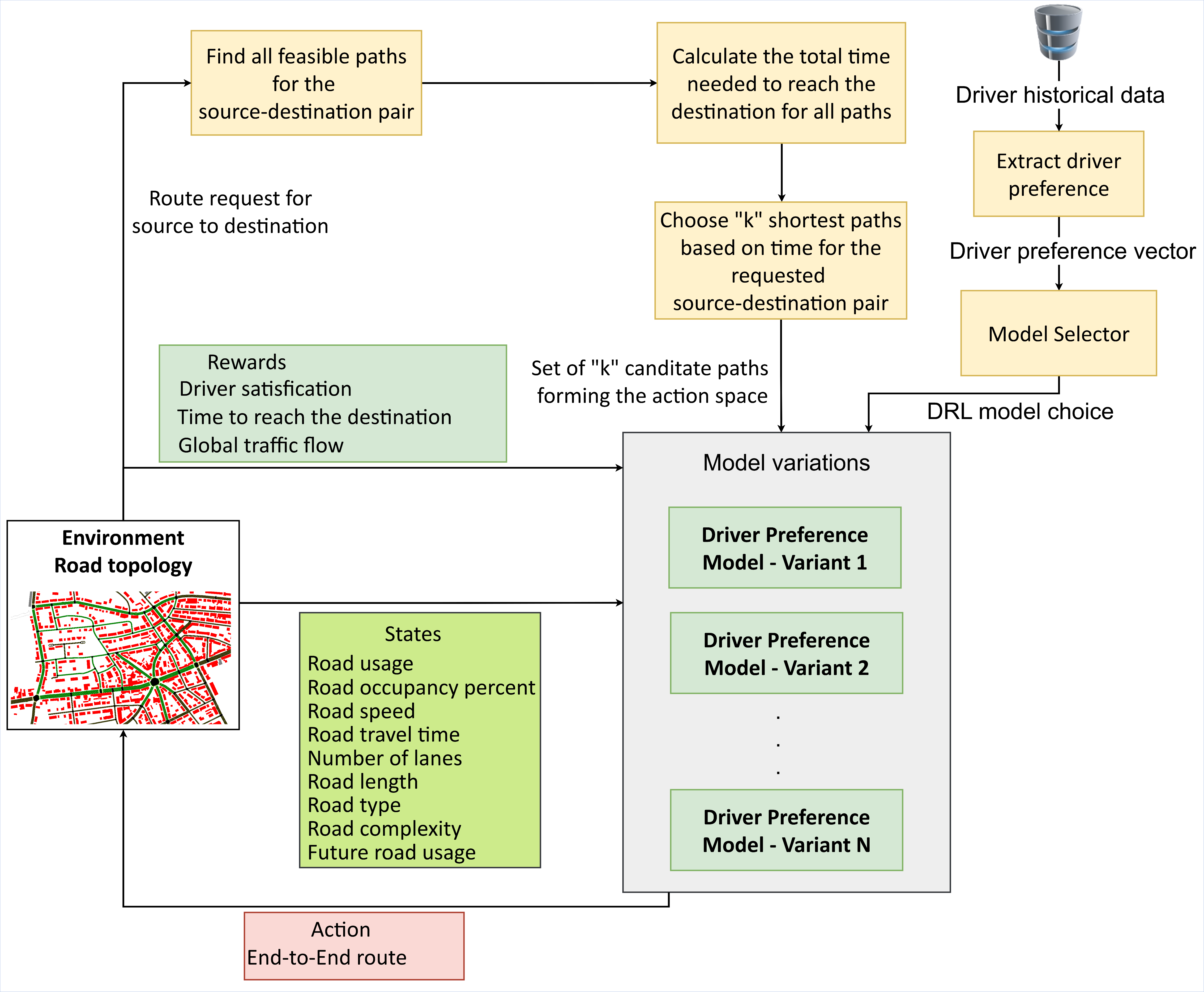}
    \caption{An overview of the proposed GNN-based DRL framework.\label{fig:flowChart}}
\end{figure}

\textbf{\emph{Learning process:}}
The learning process comprises two steps. 
First, a generic GNN-based DRL model is trained with global traffic flow and time-to-reach-destination reward components, focusing solely on traffic route optimization.
The rationale behind this choice is to establish a generic model that addresses the core process of route optimization, namely, traffic route optimization.
\textcolor{black}{Second, to accommodate personalized and context-aware factors, the generic model is further trained to consider the driver's satisfaction reward component, referred to as driver preference models.}
This two-step learning process provides us with the flexibility to incorporate different variations of driver's  preference vectors $\Pv$ on top of the basic routing functionalities provided by the generic model.
Since multiple driver preference models are trained, an identifier is necessary to select the appropriate DRL model as per the requesting driver's preferences. 
Hence, during the inference phase, we employ a cosine similarity scoring function to map trained models to driver's preferences. 
Then  we select a model with an environment attribute vector $\mathbf{h}_v$ that is similar to the driver's preferences $\mathbf{p}_v$, and the selected model is used to determine the most suitable routes for the driver.
\Cref{fig:flowChart} outlines  the proposed GNN-based DRL framework.


The framework constantly monitors real-time data from the vehicle and the current traffic congestion status to change the route if necessary.
The suggested route is re-planned if any of the three following conditions are satisfied:
(i) Driver deviates from the suggested route;
(ii) Real-time estimation of driver's behavior diverges from the behavior previously associated with the preference vector;
(iii) The current route's traffic condition has changed, causing a significant delay in the estimated time of arrival compared to the ideal time.

%

\section{Performance Evaluation\label{sec:peval}}

We evaluated the performance of the proposed framework using a real-world road network with mobility traces sampled from the Luxembourg SUMO Traffic (LuST) Scenario \cite{codeca2017luxembourg}.
We extracted a portion of the Luxembourg road network that comprises 44 intersections and 122 roads, with 5 intersections containing traffic lights.
\textcolor{black}{We selected an 8:00am morning rush hour scenario for training, while the inference phase utilized data from the 1:00pm and 6:00pm datasets, corresponding to the afternoon and evening rush hours, respectively.}



\subsection{Model pre-training}

Initially, the generic model is trained solely using time to reach the destination and global traffic flow reward components to find the optimal route based on time and congestion levels.
Once the generic model achieves satisfactory and consistent rewards, the best-performing model is selected and further trained to incorporate driver's satisfaction rewards to determine personalized routes.
In both steps, we run the GNN-based DRL framework for 150 steps per simulation episode, meaning that 150 vehicles are introduced into the road network with source and destination nodes randomly sampled from the road network.
The training process is repeated until stable and reasonable rewards are achieved.
Additionally, we freeze the learning process every 30 episodes during training, and conduct five evaluation episodes to assess the generic agent's performance.
In both pre-training and training concerning driving preferences, the number of candidate paths ($k$) is set at $4$.
During training, the DQN utilizes an $\epsilon$-greedy exploration process to encourage the agent to explore the action space and avoid local minima.
The $\epsilon$ value gradually decreases as the training progresses.



\begin{figure}
    \centering
    \vspace{-1em}
    \includegraphics[width=\columnwidth]{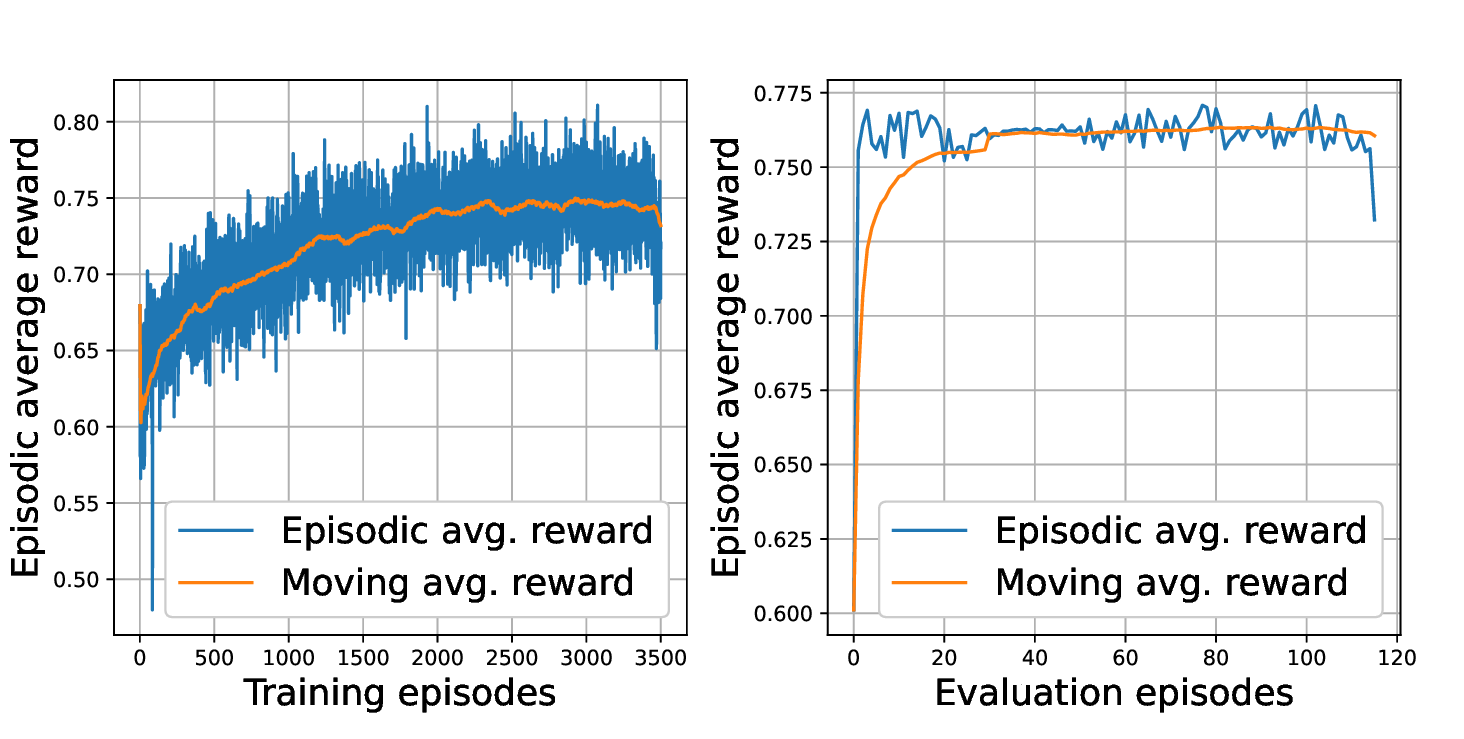}
    \vspace{-1em}
    \caption{GNN-DRL learning progress: Mean reward trend.\label{fig:trainProgress}}
\end{figure} 

As depicted in \cref{fig:trainProgress}, training process comprises of 3,500 episodes with weighting coefficients $\omega_{p}{=}0, \omega_{t}{=}0.7, \omega_{f}{=}0.3$ corresponding to the reward components defined in \cref{eq:reward}.
During training, the generic agent achieved an average episodic reward of 0.7. 
The fluctuations in the training rewards indicate that the $\epsilon$-greedy exploration effectively explores the action space, while the evaluation episode rewards are relatively stable.
\textcolor{black}{
During the inference phase, we chose the best model and ran 5 episodes with different random seeds, resulting in the agent achieving average rewards of 0.76 and 0.77 for the afternoon and evening traffic data, respectively.
As a performance metric, we calculated the median difference between the expected and ideal travel time ($T_i$) from the source to the destination, which is $24$\,s (afternoon) and $23$\,s (evening). 
Furthermore, compared to the shortest distance-based approach, our framework show an improvement of $33$\% (afternoon) and $46$\% (evening) in the mean time difference.
Additionally, the cumulative distribution function (not included in the paper due to space constraints) underlines that, on average between afternoon and evening data, the time difference is less than $22$\,s for $80$\% of instances in case of the generic model, whereas it is $46$\,s for the distance-based approach, thus  indicating longer waiting time.
}



\begin{figure}
    \centering
    \includegraphics[width=1\columnwidth]{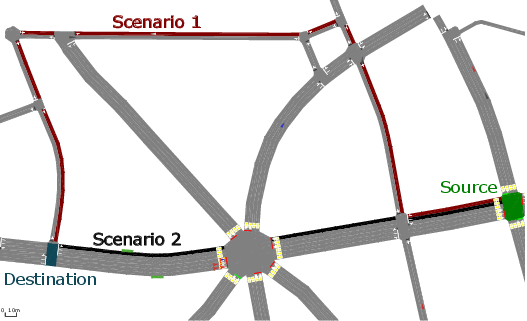}
    \caption{Route selection considering traffic light phases.\label{fig:routeComp}}
\end{figure}

To further evaluate the chosen model, we selected a specific source and destination node for a routing request, with a traffic light positioned between them. 
In the first scenario (\cref{fig:routeComp} labeled as scenario 1), the traffic light displays a red signal, and it is estimated to take another 120 seconds for the light to turn green.
Therefore, instead of opting for a direct path, an alternative path with a shorter estimated travel time is chosen to avoid the waiting time.
In the second scenario (\cref{fig:routeComp} labeled as scenario 2), although the traffic light is still red, the estimated time for the green light is less than 20\,s.
Consequently, a shorter path is chosen compared to the first scenario.
These outcomes demonstrate the advantage of incorporating dynamic traffic-related features into the framework to select the most suitable path between the source and destination.

\subsection{Training driver's preferences}

\begin{figure}
    \centering
    \vspace{-1em}
    \includegraphics[width=\columnwidth]{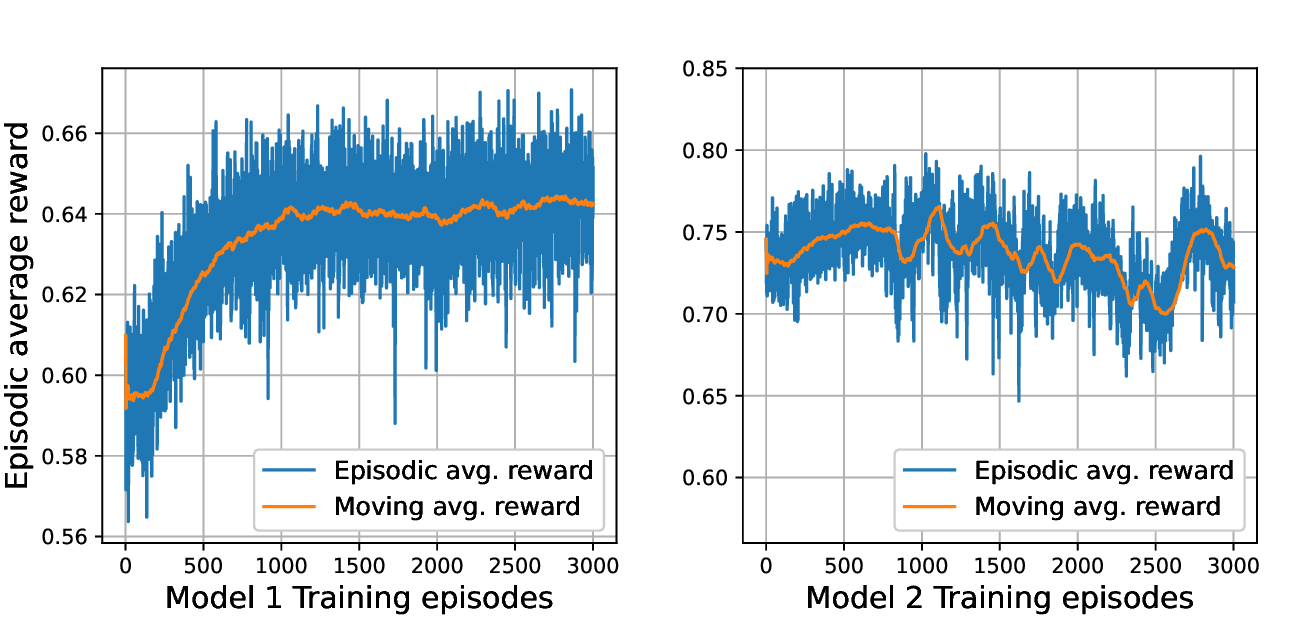}
    \caption{GNN-DRL learning progress: Mean reward trend with driver preferences. $DPM-V1$ (Left) considers [straight, two, simple, low] as the preference vector, $DPM-V2$ (Left) considers [straight, one, simple, low] as the preference vector. \label{fig:trainProgress_MM}}
\end{figure}

As the second step in the training process, the selected generic model is further trained to accommodate driver preferences with weighting coefficients $\omega_{p}{=}0.6, \omega_{t}{=}0.3, \omega_{f}{=}0.1$, accounting for driver's satisfaction, time, and congestion reward, respectively.
\textcolor{black}{In this case, we trained two driver preference-based models  (in short $DPM$)  with variations of the driver preference vector ${\Pv}$ representing road type, number of lanes, road complexity, and traffic condition:
(i) $[\textit{straight, two, simple, low}]$, and 
(ii) $[\textit{straight, one, simple, low}]$.}
The first variation, represented as DPM-V1, prefers straight roads with two lanes, simple intersections, and low traffic, while the second variation (DPM-V2) opts for single-lane over two-lane roads.
Additionally, the driver's  satisfaction reward is designed to prioritize road type and the number of lanes, while road complexity and traffic condition were given lower priority.
For the training process, training was conducted for 3,000 episodes with evaluations every 30 episodes, similar to the first case.
\Cref{fig:trainProgress_MM} shows the training progress of the second step.
\textcolor{black}{As in the first case, the best model was selected based on the  reward value during the inference phase, focusing on the evening rush hour.}

\begin{figure}
\centering
    \includegraphics[width=\columnwidth]{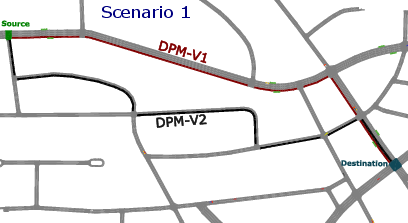}
     \includegraphics[width=\columnwidth]{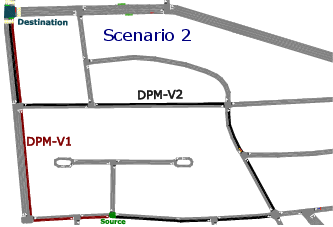}
    \caption{Route selection for reaching the destination while considering driver preferences, demonstrated through two scenarios. In both scenarios, DPM-V1 opts for the brown path, while DPM-V2 selects the black path.
    \label{fig:MMrouteComp}}    
\end{figure}

\textcolor{black}{During the inference phase, DPM-V1 and DPM-V2 achieve mean driver's  satisfaction rewards of $0.58$ and $0.80$, respectively. 
In comparison, routes chosen by DPM-V1 and DPM-V2 contain edges with similarity value greater than 0.5, ($S_e{>}0.5$) 21\%  of the time and 72\% of the time (resp.). 
The lower similarity reward of DPM-V1 is due to the fewer number of two-lane roads ($40$) compared to single-lane roads ($68$) in the road network. 
Moreover, the driver preference-based agent had to select a route from $k$ candidate paths that balance both travel time and driver satisfaction; thus, 
 it is challenging for DPM-V1 to choose a route with preferred attributes without compromising the travel time reward.
Even when the number of candidate paths ($k$) is increased from $4$ to $5$, the mean driver satisfaction reward remains at $0.58$ for DPM-V1 and $0.81$ for DPM-V2, suggesting that expanding the action space to account for more routes does not necessarily lead to better outcomes.
When compared to the generic model, DPM-V1 show a $17$\% improvement in selecting preferred routes, while DPM-V2 achieves a $7$\% improvement.
The minor improvement by DPM-V2 is because the road network attributes predominantly align with its preferences.
In terms of computation time, averaging over 750 steps, it takes about 0.16\,s to identifying a route  per source-destination request. }


To compare the routes chosen by the two models, identical source and destination pairs were selected to determine the route.
As depicted in \cref{fig:MMrouteComp}, DPM-V1 (brown path) selected the path with two lanes, whereas DPM-V2 (black path) chose the route with a single lane to reach the same destination, indicating personalized routes catering to the driver's preferences as expected.
Although this work presents only two models, multiple variations can be trained atop the generic model based on preference vectors.




%

\section{Conclusions\label{sec:conc}}

We presented a framework for a personalized and context-aware route planner that integrates driver preferences into route selection using a novel GNN-based DRL framework.
We exploited the advantages of GNNs by representing road network data as graph-structured data and employed reinforcement learning to learn the dynamics of the environment.
GNN leverages the message-passing process to share road network attributes across multiple nodes of the graph efficiently and collaborates with DRL to identify optimal routing paths, guided by reward components focused on driver's satisfaction, travel time, and congestion.
The framework offers flexibility to train multiple preference variations and assign importance to each preference factor as per the driver's wish.
We evaluated the proposed approach with a realistic road network extracted from the LuST scenario, showing that  the framework effectively learns the driver's preferences.
\textcolor{black}{
Specifically, the driver's preferences models showed up to $17$\% improvement in selecting routes aligned with driver's preferences compared to the generic model, which focuses on traditional traffic factors, and a decrease in the travel time by up to $46$\%  relatively  to the shortest distance-based approach.
Future work will focus on improving the action space of the DRL to enhance the route selection process and the driver's preference vector  to encompass additional relevant road attributes. Additionally, larger road networks will be employed to assess the computational efficiency and scalability of the proposed framework.
}

%

\section*{Acknowledgments}
This work was supported in part by the European Union's Horizon 2020 Research and Innovation Programme under Grant No.\,101069688 (CONNECT) and   by MAAS4Italy Programme (PNC-A.1-N1) under ToMove Project. The findings reported reflect the work, and are solely under the responsibility of the authors.



\end{document}